\documentclass{article}
\usepackage[T1]{fontenc}
\usepackage[utf8]{inputenc}
\usepackage[english]{babel}
\usepackage[font=small,labelfont=bf]{caption}

\usepackage{floatrow}
\newfloatcommand{capbtabbox}{table}[][\FBwidth]

\usepackage{blindtext}

\usepackage[final]{corl_2019} 
\usepackage[utf8]{inputenc} 
\usepackage[T1]{fontenc}    
\usepackage{hyperref}       
\usepackage{url}            
\usepackage{booktabs}       
\usepackage{amsfonts}       
\usepackage{nicefrac}       
\usepackage{microtype}      
\usepackage{graphicx}
\usepackage{amsmath}
\usepackage[symbol]{footmisc}

\title{Neural Embedding for Physical Manipulations}

\newcommand*\samethanks[1][\value{footnote}]{\footnotemark[#1]}
\author{Lingzhi Zhang \textsuperscript{1}\thanks{Indicates equal contribution.}  \qquad Andong Cao \textsuperscript{2}\samethanks \qquad Rui Li\textsuperscript{1} \qquad Jianbo Shi\textsuperscript{1}  \\
\textsuperscript{1}University of Pennsylvania \hspace{0.5cm} \textsuperscript{2}Yale University\\
{\tt\small \{zlz, lirui613, jshi\}@seas.upenn.edu, antonio.cao@yale.edu}
}

\begin{document}
\maketitle



\begin{abstract}
In common real-world robotic operations, action and state spaces can be vast and sometimes unknown, and observations are often relatively sparse. How do we learn the full topology of action and state spaces when given only few and sparse observations? Inspired by the properties of grid cells in mammalian brains, we build a generative model that enforces a normalized pairwise distance constraint between the latent space and output space to achieve data-efficient discovery of output spaces. This method achieves substantially better results than prior generative models, such as Generative Adversarial Networks (GANs) and Variational Auto-Encoders (VAEs). Prior models have the common issue of mode collapse and thus fail to explore the full topology of output space. We demonstrate the effectiveness of our model on various datasets both qualitatively and quantitatively. 
\end{abstract}


\begin{figure}[h]
    \centering
    \includegraphics[width=0.65 \textwidth]{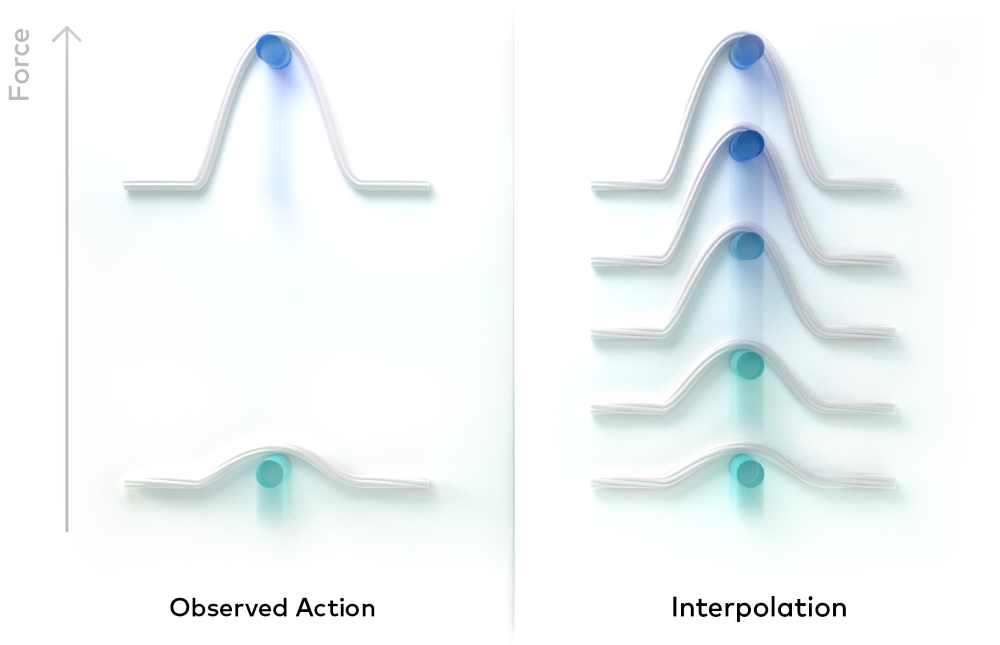}
    \caption{Given a set of sparse observations of action and state, we aim to learn a generative model that can interpolate the intermediate actions and predicts the corresponding future states. }
    \label{synthetic_demo}
\end{figure}

\section{Introduction}

Grid cells, the grid-like neural circuit in mammalian brains, is known to dynamically map the external environment as the animal navigates the world \cite{moser2015place}. Remarkably, this encoding preserves metric distance relationships, such that objects close in the real-world are close in the brain’s intrinsic map \cite{epstein2017cognitive}. Moreover, with a few observations and actions, the grid cells can rescale the mapping according to changes in the size and shape of the environment \cite{barry2007experience}. Such mental model allows quick adaptation to new surroundings, efficient localization and path-planning, and imagination of unseen events.

Inspired by the properties of grid cells, we propose a novel constraint on the latent space of a generative model that achieves data-efficient discovery of output spaces. Similar to the grid cell’s distance-preserving encoding of the world, our proposed model preserves the normalized pairwise distance of samples between the parametric low-dimensional latent space and high dimensional output space. Intuitively, this approach encourages the neural network to actively explore the action space and “stretches” the latent space outward, thus making the learned embedding as diverse as possible. Such property enables the model to interpolate and decode to unseen states, mimicking the brain’s ability to make accurate interpolations given sparse examples. 

In real-world robotics operations, common tasks like predicting ball collision or rope manipulations can involve extremely vast action and state spaces. But often we only see relatively sparse observations. A common approach to encode such high-dimensional spaces is to use deep generative models such as Generative Adversarial Networks (GANs) and Variational Auto-Encoders (VAEs). However, these generative models suffer from mode collapsing and mode dropping, where the models can only capture a partial real data distribution. These symptoms are especially problematic for our application, where the goal is to explore and represent the full unknown topological structure of the action and state spaces.



We propose a method that effectively solves the problem of mode collapsing by learning a distance-preserving mapping from the latent space to the output space. Moreover, our model is trained with adversarial learning that enforces the generated samples to be plausible. These properties enable the model to learn a latent encoding of a given task with a few samples, and interpolate based on the learned encoding to predict future events. 

Our model has several practical applications: first, during many robotic operations, the constraints on the action space, such as safety or geometric limits specific to the task, are dynamic and largely unknown, whereas a wide range of possible maneuvers remains uncharted. Our proposed method can help the robot efficiently explore the action space and predict the future states. Second, the latent space in our model supports safe interpolation in the output space, thus enabling robots to output reliable and plausible action proposals. Third, our model can accurately predict future states given current states and sampled actions. Therefore, this method can help a robot build a physical mental mapping of the task at hand.



The main contributions of this work can be summarized as follows: 
\begin{itemize}
  \item We propose a generative model that mimics the idea of grid cell that can approximate the unknown action/state space with only sparse observations.  
  \item We generate a synthetic dataset and two simulation datasets, and have demonstrated our model can outperform strong baseline generative models on these datasets. 
\end{itemize}

\section{Related Work}
Recent works have made substantial progress in imposing diversity constraint on the latent space of a generative model. In particular, Liu et. al. \cite{ndiv} proposes the normalized diversification technique that effectively solves the problem of mode collapsing. Building on top of their prior work, we use a similar technique to learn an accurate encoding of action and state spaces in physical manipulation tasks. To our knowledge, our model is the first to use normalized diversification in these applications.

\textbf{Action-conditioned dynamics learning:} Action-conditioned dynamics learning \cite{oh_action, finn_physical, jiajun_wu_deanim, watter_dynamics} aims to predict future states given current states and actions. For instance, Finn et. al. \cite{finn_physical} introduces an action-conditioned model that can predict object motion without labels, and Wu et. al. \cite{jiajun_wu_deanim} proposes a paradigm for visually de-animate a scene, and make future predictions using generative models. Our proposed approach extends these prior works by introducing a generative model that samples actions given current state and then make predictions, while allowing safe interpolation on the latent action space.

\textbf{Exploration:} Our work has some relation to exploration in reinforcement learning. Relevant strategies to explore the action and state space fall under two categories: Intrinsic motivation models and representation learning models.

 Intrinsic motivation models \cite{stadie_explore, Houthooft_curiosity, Bellemare_motivation, fu_exploration, achiam_sastry_explore, pathak_curiosity, Ostrovski_explore} typically define an intrinsic reward that encourages the agent to explore unseen states and actions. For example, Pathak et. al. \cite{pathak_curiosity} uses the error of the agent’s prediction of the consequence of its action as the intrinsic reward, and Stadie et. al. \cite{stadie_explore} evaluates the learned model of the system dynamics to assign bonus to the agent. A common choice of an intrinsic reward is information-theoretic errors, such as the mutual information between the agent’s states and actions \cite{gregor_intrinsic_control, Houthooft_curiosity, ziebart_max_ent, haarnoja_deep_energy, haarnoja_max_entropy, Jung_empowerment, DIAYN}. Our work proposes a different approach to enforce diversity on the learned action space, and does not require any direct interaction with the system to calculate the reward.

 Representation learning models aim to explore high-level representations of the action and state spaces in order to enable interpolation and transfer of skills. A popular method is to use hierarchical policies \cite{Haarnoja_latent_hiearhical, Coros_robust_task_control,Frans_meta_hierarchy, Liu_graphics, merel_visuomotor, Peng_terrain_locomotion}, where a high-level policy selects a set of low-level skills to solve a given task. Training the hierarchical policy requires pre-training a set of primitives that can then be combined together, and the typical hierarchical policy can only select one task at a time. Another common approach in representation learning is to encode the learned skills onto a latent space \cite{Kolter_omni_07, Haarnoja_latent_hiearhical, Heess_learning_transfer, DIAYN, hausman_embedding, merel2018humanoid, peng_learning_hierarchical_control}, using generative models such as VAEs. For example, Achiam et. al. \cite{Achiam_var_option_discovery} uses a VAE to decode from trajectories in order to learn distinct and dynamical options. Although our work targets a different task, namely to encode the action and state spaces of a physical system, our proposed model presents an alternative to the prior diversity-enforcing methods, while also allowing safe interpolation in the latent space.

\textbf{Generative Models.} In recent deep learning works, there are two types of popular generative models, which are Variational Auto-Encoder (VAE) \cite{vae} and Generative Adversarial Network (GAN) \cite{gan}.  The VAE model intends to optimize a variational lower bound on the data log-likelihood, and auto-encodes the data to learn a compact latent space. On the other hand, the GAN network has a generator that maps a latent space to an output space and approximates the data distribution, and a discriminator predicts a probability of the sample coming from the real data rather than a generator. A meaningful latent space is learned once the generator and discriminator achieve an adversarial equilibrium. There are several applications that use the latent space property to synthesize image \cite{gan_app_1, gan_app_2, gan_app_3, gan_app_4} and interpret visual objects \cite{gan_app_5}.  

\section{Methodology}
\begin{figure}[h]
    \centering
    \includegraphics[scale=0.181]{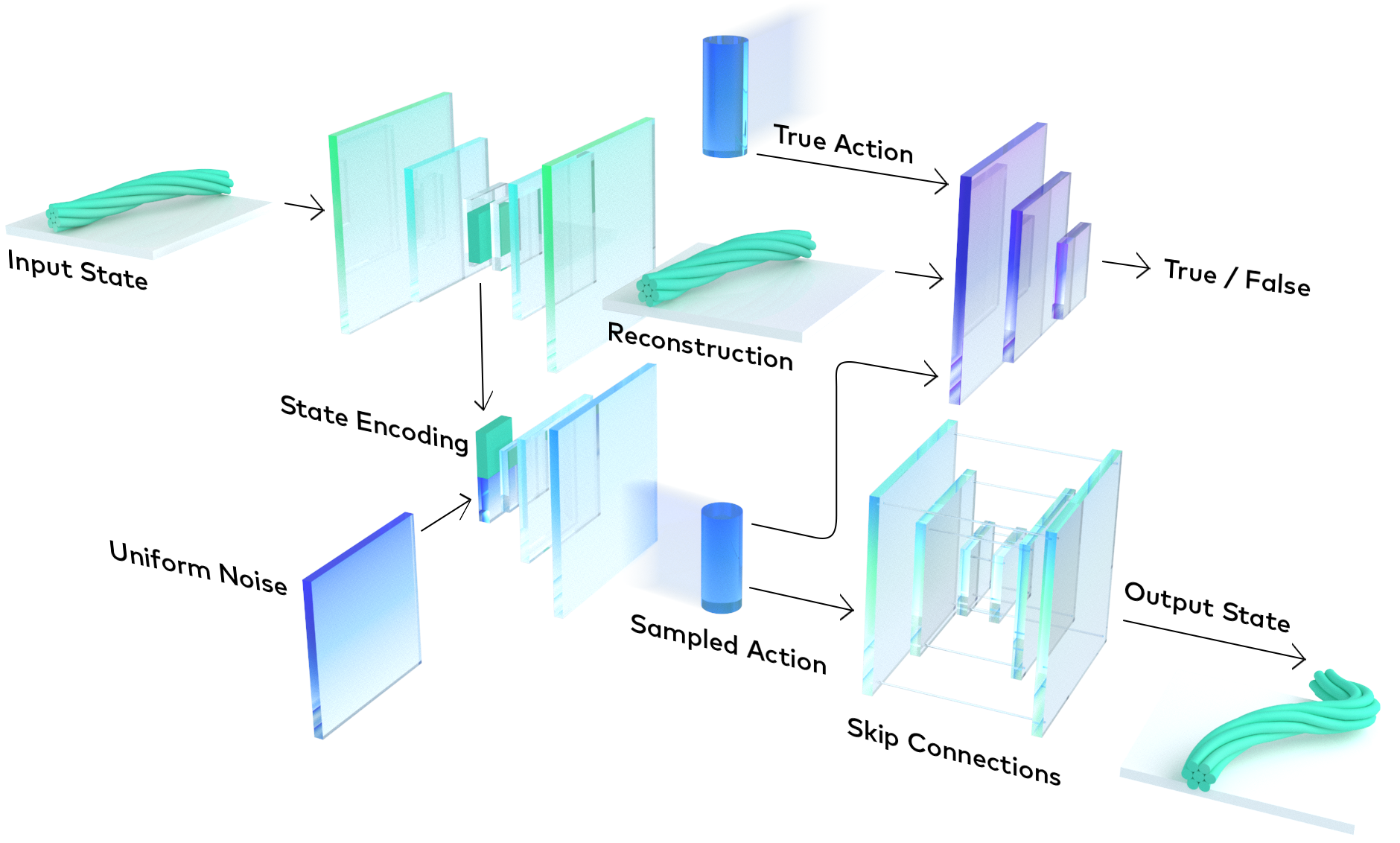}
    \caption{This is an overview of our model architecture. \textbf{Top Left:} An \textbf{auto-encoder} that guides the network to learn a meaningful feature embedding of the input state. \textbf{Bottom Left: } The \textbf{action decoder} takes the input state embedding concatenated with a noise sampled uniform distribution and predicts an action. \textbf{Top Right: } Conditioned on the input state, the \textbf{discriminator} takes actions as inputs and predicts the probability of whether the action is from training data or generator. \textbf{Bottom Right: } The \textbf{kinematics forward model} takes the input state and the action as input, and predicts a output state in a deterministic way. }
    \label{figure:architecture}
\end{figure}

In most real-world scenarios, a robot can perform many stochastic actions given a current state, and can reach a deterministic future state given a current state and an action. Thus, we consider the mapping from current state to action as a variational process, and the mapping from a pair of current state and action to the future state as a deterministic process. We learn a multimodal generative model to predict diverse actions given a current state, and a deterministic forward kinematics model to predict a future state when given a pair of current state and action. Our model is shown in figure \ref{figure:architecture}. 

\subsection{Generative Model to Unfold Action Space}

\begin{figure}[h]
    \centering
    \includegraphics[scale=0.18]{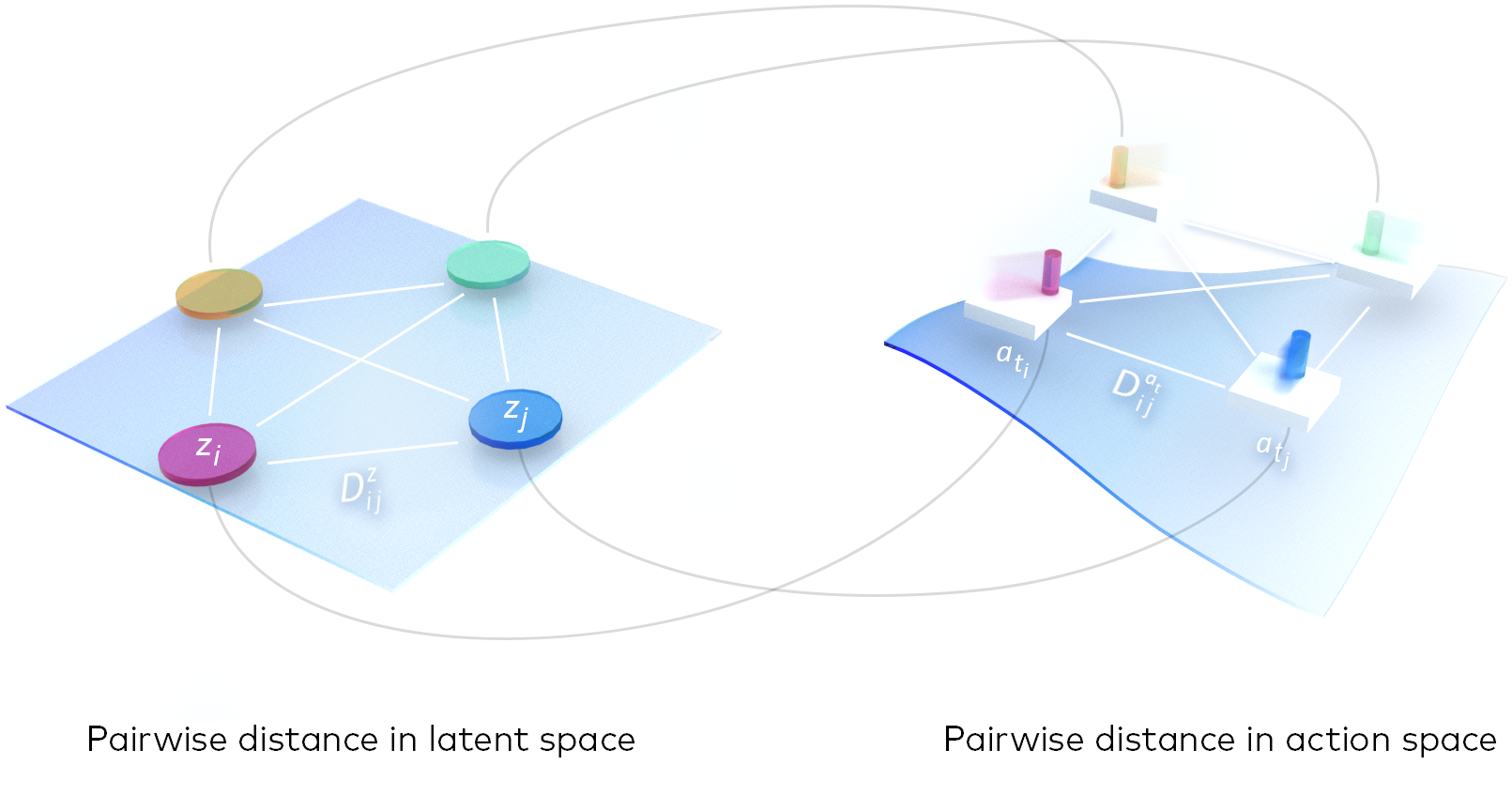}
    \caption{This figure shows the idea of normalized pairwise distance in the latent space and action space.}
    \label{figure:pairwise_distance}
\end{figure}

The first part of our method includes a generative model that approximates action space with unknown topology. Using sparse observations as our input, we train an auto-encoder that could guide the network to learn a meaningful feature embedding that encodes important state information. Such feature embedding enables the network to share knowledge across sparse observations of similar input states. Then, we concatenate the state embedding with a random latent variable sampled from an uniform distribution, and decode it to a predicted action through an action decoder. To encourage active exploration of action space, a normalized diversity loss is imposed to preserve the normalized pairwise distance between latent variables and sampled actions, as shown in figure \ref{figure:pairwise_distance}. A discriminator is co-trained to predict probability of the actions coming from either the training data or the generator, which enforces the sampled actions to be plausible.

\subsubsection{Active Exploration Via Normalized Diversification} 

When mapping random variables from the latent space to the action space, our generative model preserves the normalized pairwise distance of different generated samples in between the latent space and the action space. The distance metric $d_z(.,.)$ between any two samples is simply a Euclidean distance. We denote $z$ as latent variables, $a$ as actions, and $i,j$ as sample indices. 

\begin{equation}
    d_z(z_i, z_j) = ||z_i - z_j|| \quad \text{,} \quad
    d_{a_t}(a_t^i, a_t^j) = ||a_t^i - a_t^j||
\end{equation}

Furthermore, the normalized pairwise distance matrices are defined as $ D_{ij}^z$, $D_{ij}^{a_t}$ $\in$ $\mathbb{R}^{N \times N}$ as follows, 

\begin{equation}
    D_{ij}^z = \frac{d_z(z_i, z_j)}{\sum_j d_z(z_i, z_j)} \quad \text{,} \quad
    D_{ij}^{a_t} = \frac{d_{a_t}({a_t}_i, {a_t}_j)}{\sum_j d_{a_t}({a_t}_i, {a_t}_j)}
\end{equation}

During training, we treat the normalizer in (2) as a constant when back-propagating the gradient to the generator network. This ensures that we optimize the absolute pairwise distance for a sample, rather than adjusting normalizer to satisfy the loss constraint. The normalized diversity loss function is defined as follows, 

\begin{equation}
    \mathcal{L}_{ndiv} (s_t, a_t, z) = \frac{1}{N^2-N}\sum_{i=1}^N\sum_{i \ne j}^N max(0, \alpha D_{ij}^z - D_{ij}^{a_t})
\end{equation}

where $\alpha$ is a hyperparameter. We do not consider the diagonal elements of the distance matrix, which are all zeros.

Unlike GANs and VAEs, our generative model parameterizes the latent space as a uniform distribution ${U}(0,1)$ instead of Gaussian distribution. There are two reasons. First, the uniform distribution is bounded so that the sampled latent variables will never be too far away from each other. Sampling latent variables too far away from each other might induce extremely large pairwise distance and thus might lead to exploding gradients when optimizing the loss. Second, the Gaussian distribution makes a strong assumption that the data has a mode to fit the distribution, while uniform distribution has the flexibility to map to diverse modes in the data distribution. A simple way to think of this is to cut the uniform distribution space into many different pieces and learn mapping for each piece to fit each mode in the data distribution.

\subsubsection{Safe Mapping Via Adversarial Training}

While the normalized diversity loss encourages the model to actively explore in the action space, the adversarial loss puts a constraint during exploration so that the predicted actions are plausible. Our adversarial training framework is based on conditional GAN \cite{conditional_gan}. The action decoder takes an input state encoding, concatenates the encoding with a random variable sampled from a ${U}(0,1)$ latent space, and finally decodes to a predicted action. The discriminator takes both real and generated actions as inputs and predicts whether the action is real or fake conditioned on the input state. During implementation, we use the concatenation of the action and input state embedding as the input of the discriminator, and we use hinge loss \cite{geometricgan} \cite{trandeep} to train the generator and discriminator,

\begin{equation}
    \mathcal{L}_{D} (s_t, a_t, z) = \mathbb{E}_{a_t \sim q_{data}(a_t)}[ \mbox{min}(0, 1-D(a_t|s_t))] + \mathbb{E}_{z \sim p_(z)}[\mbox{min}(0, 1+D(G(s_t, z)|s_t))]
\end{equation}

\begin{equation}
    \mathcal{L}_{G} (s_t, z) = - \mathbb{E}_{z \sim p_(z)}[D(G(s_t, z)|s_t)] \quad \text{,} \quad
    \mathcal{L}_{adv} = \mathcal{L}_{D} + \mathcal{L}_{G}
\end{equation}

To stabilize training, spectral normalization \cite{sngan} is applied to scale down the weight matrices in discriminator by their largest singular values, which effectively restricts the Lipschitz constant of the network. The generator and discriminator are updated alternatively in each iteration. After training converges to an equilibrium, the generator is able to sample diverse and plausible actions given a current state. 

\subsection{Forward Kinematics Model to Predict the Future State}

In the second part of our method, we use a forward kinematics model to predict future states by inputting a pair of current state and action. We consider predicting a future state as a deterministic process, and thus we train a standard network to regress the predicted future state towards the ground truth future state with a Euclidean reconstruction loss function as follows, 

\begin{equation}
    \mathcal{L}_{recon} (s_{t+1}, {s_{t+1}}^*) = ||s_{t+1} - {s_{t+1}}^*||
\end{equation}

where $s_{t+1}$ and $s_{t+1}^*$ are ground truth and the predicted future states respectively. The states could be high-dimensional images or some low-dimensional parameterizations depending on different applications.

\section{Experimental Results}

\subsection{Preliminaries}
We conduct experiments on one synthetic dataset and two simulation datasets. Both simulation datasets are generated on the Unity game engine.

The first simulation dataset (shown in the left-hand-side of Figure 4) contains an orange capsule and a ground plane both with fixed friction coefficients. For each data, a point on the capsule’s waist is sampled, and an impulse with random direction and magnitude is applied onto the sampled point on the capsule. The dataset contains the images of the capsule before being hit, and 2 seconds after being hit. The wait time is selected empirically to ensure that the capsule does not disappear from the view of the camera when the second picture is taken.

The second simulation dataset (shown in the right-hand-side of Figure 4) contains a deformable rope object, a cylinder for pushing the rope, and a ground plane all with fixed friction coefficients. For each data, a node $\vec{n}$ on the rope and a point $\vec{p}$ are randomly sampled, such that $|\vec{p} - \vec{n}| \in \{2r, 1\}$ where $r$ is the radius of the cylinder. Then a magnitude and direction is randomly sampled, and the cylinder will move along the sampled displacement with a fixed velocity. 

\begin{figure}[h]
    \label{Simulations}
    \centering
    \includegraphics[width=\textwidth]{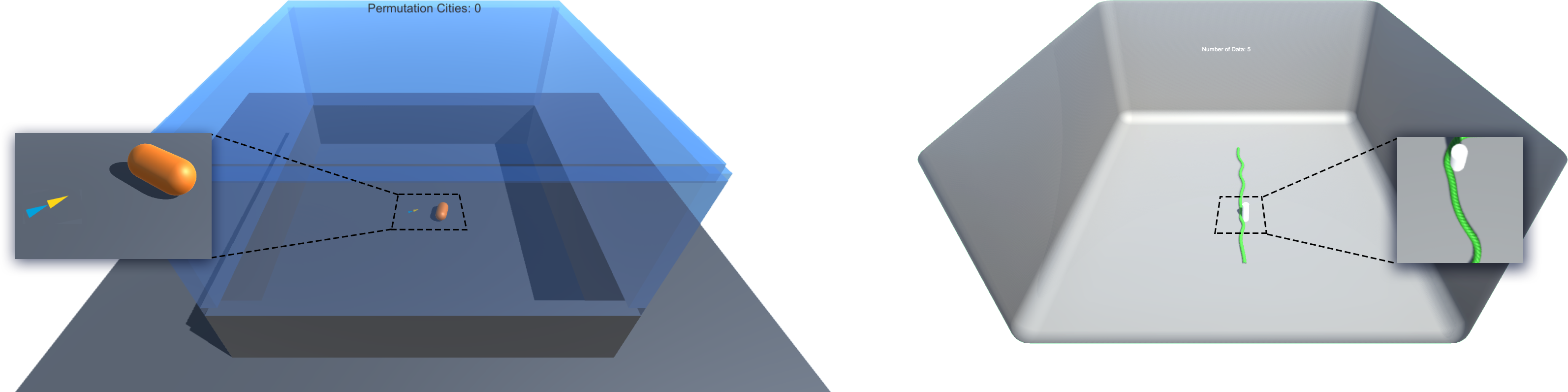}
    \caption{\textbf{Left:} Rolling dataset; \textbf{Right:} Rope dataset.}
\end{figure}

\textbf{Evaluation Metric. }
To evaluate whether the sampled actions are plausible or realistic, we use three evaluation metrics to quantify the similarity between the generated and real action distributions, including Fréchet Distance \cite{FD} and Jensen-Shannon Divergence (JS Divergence) \cite{JSD}.

\textbf{Baseline Models.} We conduct experiments in two settings. One is with a fixed initial state, and another is with various initial states. We use GAN \cite{gan} and VAE \cite{vae} as the baseline models for the first case, and conditional GAN \cite{cgan} and conditional VAE \cite{cvae} as baseline models for second case. Specifically, we use spectral normalization \cite{sngan} to stabilize GAN training. 

\subsection{Evaluation of Unfolding Action Space}

In the synthetic experiment, we model pushing a ball away from the center of a surface where the action space and state space are all unknown. The initial state is considered to be fixed at the center. We design the action space to be a star-like space that models the geometric constraint in the surface environment, and the state space to be a non-linear transformation of the action space that models irregular frictions or slopes on the surface. As shown in figure \ref{synthetic_demo}, the actions are denoted as blue dots and the states are denoted as green dots. The dots in the action space represents the orientation and magnitude of pushing force and the dots in the state space represent the location of objects after action applied. 

\begin{figure}[h]
    \centering
    \includegraphics[width=\textwidth]{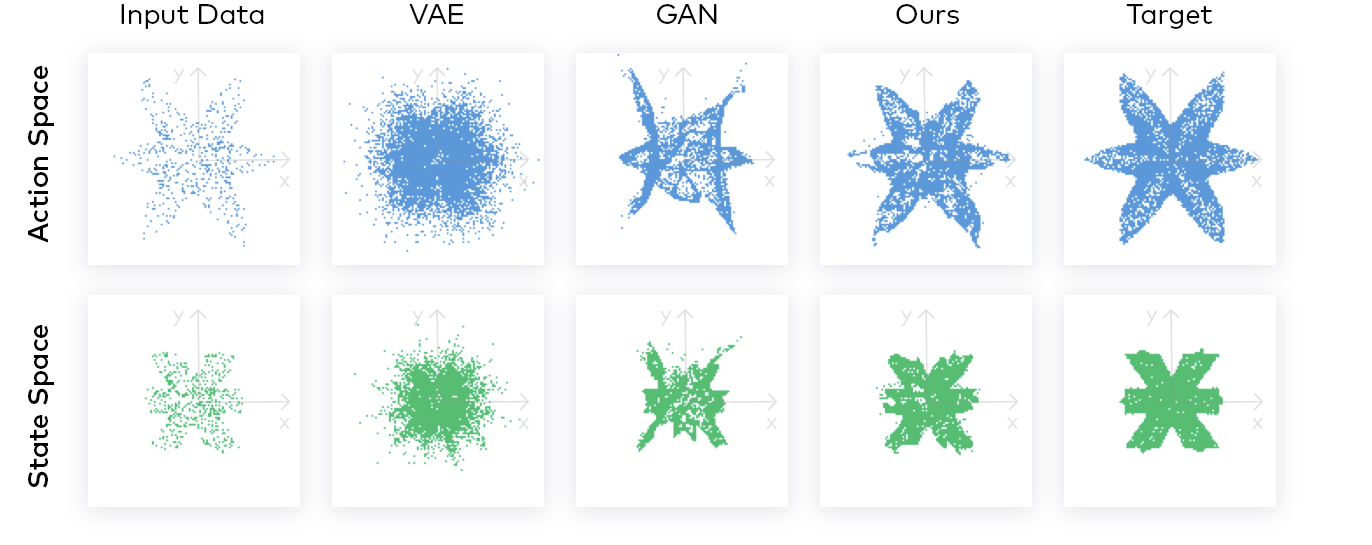}
    \caption{Comparison of generative models' ability to discover the unknown action and state spaces. }
    \label{synthetic_demo}
\end{figure}

\begin{minipage}{0.9\textwidth}
\centering
 \begin{minipage}{0.4\textwidth}
    \centering
    \includegraphics[scale = 0.35]{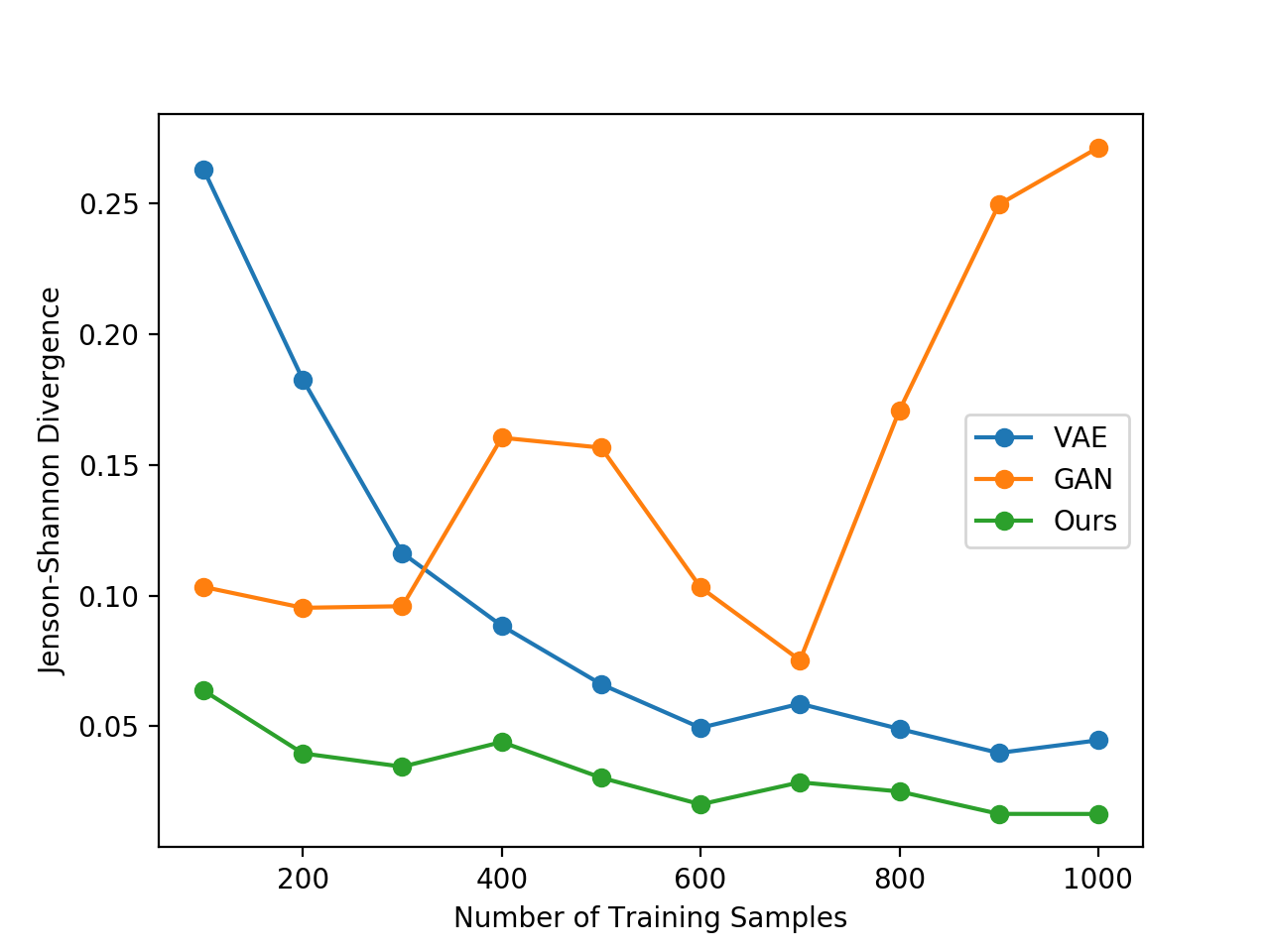}
\end{minipage}
\hfill
 \begin{minipage}[b]{0.49\textwidth}
   \begin{tabular}{c|c|c}
\toprule
 Model & Fréchet Distance $\downarrow$ & JS Divergence $\downarrow$ \\
\hline
 VAE\cite{vae}  & 21.55 $\pm$ 2.210 & 0.05 $\pm$ 0.002\\
 GAN\cite{gan}  & 26.83 $\pm$ 19.40 & 0.16 $\pm$ 0.031\\
 Ours  & \textbf{3.48 $\pm$ 0.748} & \textbf{0.02 $\pm$ 0.001}\\
\bottomrule
\end{tabular}
\end{minipage}\\
\begin{minipage}[t]{0.4\textwidth}
    \captionof{figure}{A table shows "JS Divergence" between approximate and real action distribution" versus "number of training samples".}
 \end{minipage}%
 \hfill
\begin{minipage}[t]{0.49\textwidth}
\captionof{table}{A table shows the comparison results between other generative models and ours in synthetic dataset.}
\end{minipage}%
\end{minipage}

\vspace{-8 pt}

First, we conduct a comparison study on the synthetic dataset, we train VAE \cite{vae} by encoding and decoding the action to learn a compact latent space, and train GAN \cite{gan} by directly mapping the latent space to the action space. As shown in figure \ref{synthetic_demo}, VAE is not able to capture the complex structure of the real action space with a Gaussian prior on the action space. In addition, GAN has encountered the problem of mode collapse, which means many latent variables are mapped to the same point in the action space. Superior to both baseline methods, our model is able to actively explore and safely interpolate the full action state and approximates its topological structure. For all models, we train a same forward kinematics model to map the action space to the state space. In the experiments, we use 600 points as training and sample 10,000 points to approximate the action space after training. The quantitative evaluation also indicates our model can better approximate the unknown action space, as shown in table 2. 

Second, we conduct evaluation on the rope and roller datasets. In figure \ref{synthetic_demo}, we demonstrate that our model can sample diverse and plausible actions given an input state. In table \ref{table:rope_roller_eval}, we show our model can outperform VAE\cite{vae} and GAN\cite{gan} on both datasets using Fréchet Distance \cite{FD} and Jensen-Shannon Divergence \cite{JSD}.

\vspace{-8 pt}

\begin{figure}[h]
    \centering
    \includegraphics[scale=0.3]{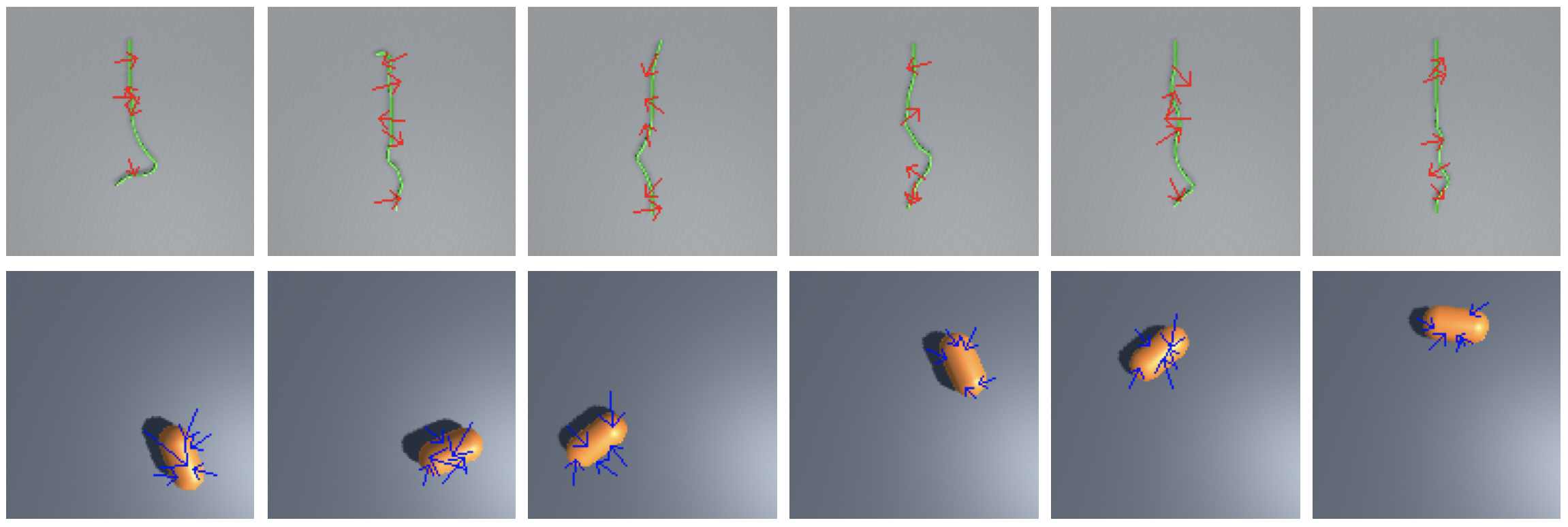}
    \caption{Quanlitative results of diverse action sampling on rope and roller manipulations. }
    \label{synthetic_demo}
\end{figure}
\vspace{-8 pt}
\begin{table}[!htb]
\begin{tabular}{c|cc|cc}
\toprule
  & \multicolumn{2}{c|}{Rope} &  \multicolumn{2}{c}{Roller} \\
\midrule
 Model & Fréchet Distance $\downarrow$  & JS Divergence $\downarrow$ & Fréchet Distance $\downarrow$ & JS Divergence $\downarrow$\\
\hline
 VAE\cite{vae}  & 12.367 $\pm$ 1.049 & 0.670 $\pm$ 0.009 & 10.140 $\pm$ 0.002 & 0.660 $\pm$ 0.006\\
 GAN\cite{gan}  & 16.481 $\pm$ 10.450 & 0.667 $\pm$ 0.007 & 13.045 $\pm$ 6.798 & 0.666 $\pm$ 0.005\\
 Ours  & \textbf{11.084 $\pm$ 4.460} & \textbf{0.547 $\pm$ 0.101} & \textbf{9.662 $\pm$ 4.905} & \textbf{0.504 $\pm$ 0.085}\\
\bottomrule
\end{tabular}
\caption{A table shows the comparison results between other generative models and ours in the rope and roller datasets. }
\label{table:rope_roller_eval}
\end{table}

\vspace{-8 pt}

\subsection{Evaluation of Future State Prediction}

The state space could be low-dimensional vectors or high-dimensional images depending on different applications. We evaluate the performance of future state prediction both qualitatively and quantitative as shown in figure and table below. With simple MSE reconstruction loss, the forward kinematics model can produce very good predictions of future states.

\begin{minipage}{\textwidth}
\centering
  \begin{minipage}[b]{0.4\textwidth}
    \includegraphics[scale = 0.4]{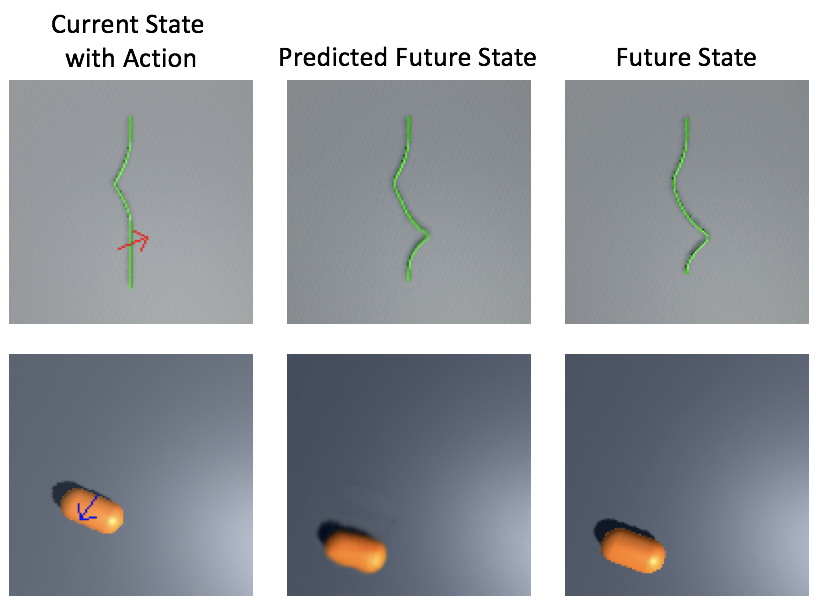}
    \captionof{figure}{Predicted and ground truth future states given input state and action. }
  \end{minipage}
  \label{figure:state_predict}
  \begin{minipage}[b]{0.49\textwidth}
  \centering
  \begin{tabular}{c|c}
\toprule
 Model & Pixel MSE Error \\
\hline
 Rope  & 5.8908 \\
 \hline
 Roller  & 54.7298 \\
\bottomrule
\end{tabular}
      \captionof{table}{Pixel level reconstruction error over 1,000 test images with 128 x 128 resolution.}
    \end{minipage}
  \end{minipage}

\subsection{Visualization of Network Feature Embedding}

In the many robotic operations, the observations of action and states are often very few and sparse. Thus, we want to ensure that the network has a nice property to share knowledge across sparse observations. Similar current states should correspond to similar action space, so it is crucial that the current state feature embedding encodes the important spatial and shape information of the target object in order to cluster them. We train an auto-encoder on top of the action prediction network to achieve this property. To evaluate the state feature embedding of the learned encoder, we use t-SNE \cite{tsne} over image features to visualize the rope and roller images. Images with similar configuration appear near each other, which indicates our state encoder learns meaningful information to capture variations of the target objects. 

\begin{figure}[h]
    \centering
    \includegraphics[width=\textwidth]{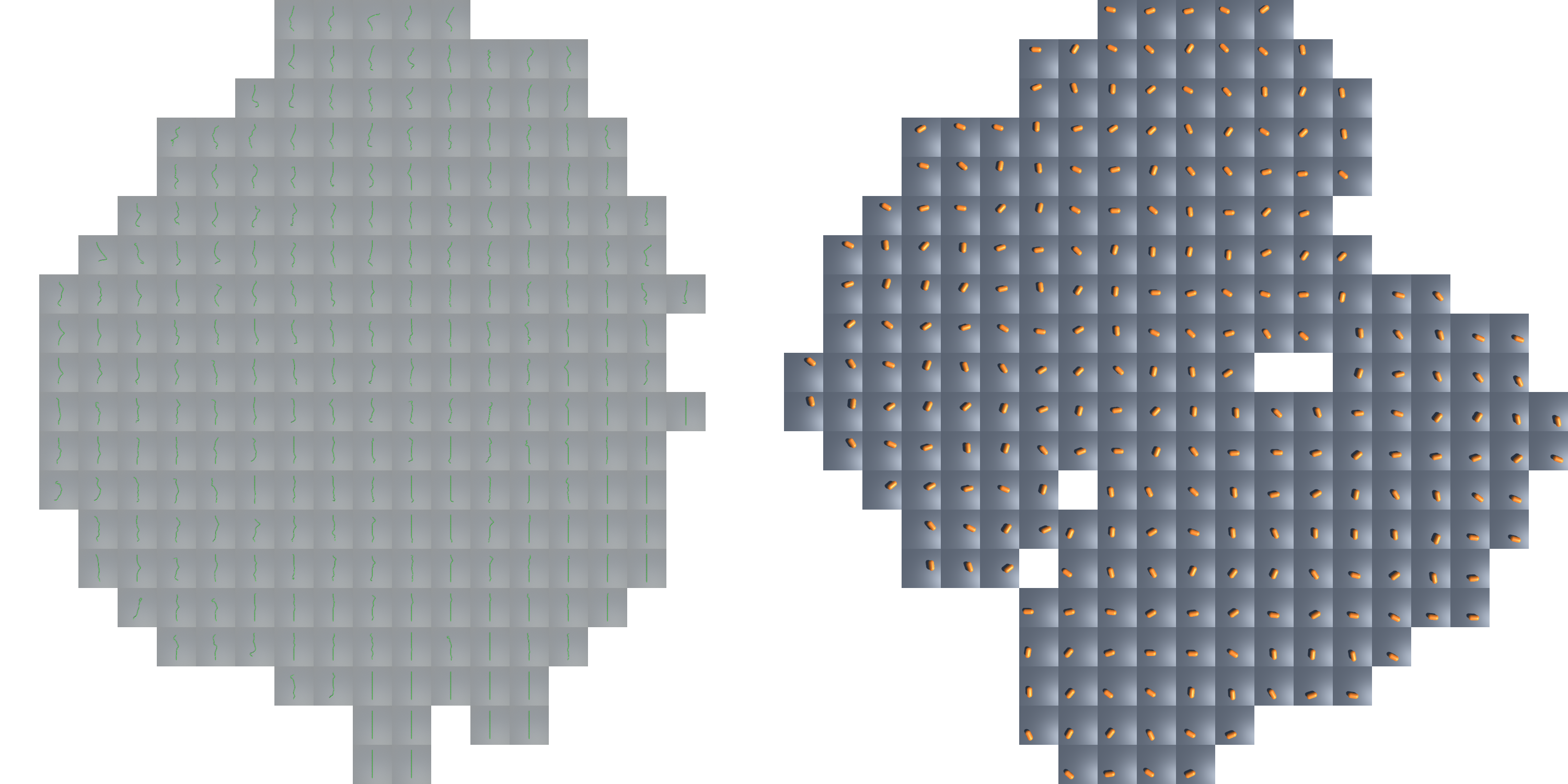}
    \caption{This figure shows the rope and roller images on the t-SNE embeddings \cite{tsne} using the feature extracted by the current state encoder. Zoom in to see the details. }
    \label{synthetic_demo}
\end{figure}

\vspace{-8 pt}

\section{Conclusion}
In this work, we propose a generative model that can approximate vast and unknown action and state spaces using only sparse observations. Current generative models suffer from mode collapsing and mode dropping issues, and so we propose a method that solves these issues by learning a distance preserving mapping from latent space and produces plausible action solutions. We generate a synthetic dataset and two simulation datasets, and have demonstrated that our model can outperform strong baseline generative models on all of these datasets. Our work proves useful for applications in robotic operations where observations of action and state spaces are limited, and in cases where full exploration of the topology is needed. Future work includes building a better exploration strategy on top of our method for reinforcement learning. 

\newpage

\bibliography{example}  

\end{document}